\documentclass{article}

\usepackage[preprint]{neurips_2024}

\usepackage[utf8]{inputenc} 
\usepackage[T1]{fontenc}    
\usepackage{hyperref}       
\usepackage{url}            
\usepackage{booktabs}       
\usepackage{amsfonts}       
\usepackage{nicefrac}       
\usepackage{microtype}      
\usepackage{xcolor}         
\usepackage{amsmath}  
\usepackage{graphicx}
\usepackage{subcaption}
\usepackage{multirow}
\usepackage{array}

\title{GKAN: Graph Kolmogorov-Arnold Networks}

\author{%
  Mehrdad Kiamari \\
  Department of Electrical and Computer Engineering\\
  University of Southern California\\
  Los Angeles, CA 90089 \\
  \texttt{kiamari@usc.edu} \\
  \And
  Mohammad Kiamari \\
  Department of Computer Science\\
  RWTH Aachen University\\
  Aachen, Germany \\
  \texttt{mohammad.kiamari@rwth-aachen.de} \\
  \And
  Bhaskar Krishnamachari \\
  Department of Electrical and Computer Engineering\\
  University of Southern California\\
  Los Angeles, CA 90089 \\
  \texttt{bkrishna@usc.edu} \\
}

\begin{document}

\maketitle

\begin{abstract}
We introduce Graph Kolmogorov-Arnold Networks (GKAN), an innovative neural network architecture that extends the principles of the recently proposed Kolmogorov-Arnold Networks (KAN) to graph-structured data. By adopting the unique characteristics of KANs, notably the use of learnable univariate functions instead of fixed linear weights, we develop a powerful model for graph-based learning tasks. Unlike traditional Graph Convolutional Networks (GCNs) that rely on a fixed convolutional architecture, GKANs implement learnable spline-based functions between layers, transforming the way information is processed across the graph structure. We present two different ways to incorporate KAN layers into GKAN: architecture 1 --- where the learnable functions are applied to input features after aggregation and architecture 2 --- where the learnable functions are applied to input features before aggregation. We evaluate GKAN empirically using a semi-supervised graph learning task on a real-world dataset (Cora). We find that architecture generally performs better. We find that GKANs achieve higher accuracy in semi-supervised learning tasks on graphs compared to the traditional GCN model. For example, when considering 100 features, GCN provides an accuracy of 53.5 while a GKAN with a comparable number of parameters gives an accuracy of 61.76; with 200 features, GCN provides an accuracy of 61.24 while a GKAN with a comparable number of parameters gives an accuracy of 67.66. We also present results on the impact of various parameters such as the number of hidden nodes, grid-size, and the polynomial-degree of the spline on the performance of GKAN.
 
\end{abstract}

\section{Introduction}

The landscape of deep learning has witnessed transformative advancements in recent years, particularly in the development of methodologies that effectively handle graph-structured data—a crucial element in applications like recommendation systems that utilize intricate user-to-item interaction and social graphs [\cite{Bronstein}, \cite{Hamilton}, \cite{Kipf}, \cite{Monti}, \cite{Rianne}, \cite{You2018GraphRNNGR}]. Among the notable innovations, Graph Convolutional Networks (GCNs) have emerged as a paradigm-shifting architecture [\cite{Hamilton}, \cite{Kipf}, \cite{Monti}, \cite{Rianne}, \cite{pinsage}]. GCNs harness the power of neural networks to iteratively aggregate and transform feature information from local graph neighborhoods [\cite{pinsage}]. This method enables a robust integration of both content and structural data from graphs, setting new benchmarks across various recommender system applications [\cite{Hamilton}]. Further improvements in accuracy are essential across many domains, including large-scale networks, to address the limitations of GCNs.

At their core, GCNs are based on Multi-layer Perceptrons (MLPs), which are foundational to modern deep learning frameworks and essential for their robust capability to approximate nonlinear functions—a trait anchored in the universal approximation theorem [\cite{MLP_universal_approx}]. Despite their widespread use and critical role in contemporary models, MLPs encounter notable limitations, such as significant consumption of non-embedding parameters in transformers [\cite{attention}] and limited interpretability unless additional post-analysis tools are employed [\cite{MLP_interpret}]. In response, \cite{KAN} have recently introduced an innovative alternative, the Kolmogorov-Arnold Networks (KAN), inspired not by the universal approximation theorem like MLPs but by the Kolmogorov-Arnold representation theorem [\cite{KA_theoren_1},\cite{KA_theoren_2},\cite{KA_theoren_3}]. Unlike MLPs, which utilize learnable weights on the edges and fixed activation functions on nodes, KANs deploy learnable univariate functions on the edges and simple summations on the nodes. Each weight in a KAN is thus a learnable one-dimensional function, shaped as a spline, allowing for significantly smaller computation graphs compared to those required by MLPs. Historically, neural network designs based on the Kolmogorov-Arnold theorem have primarily adhered to a depth-2, width-(2n + 1) model [\cite{lai2023kolmogorov_kan_2layer}], lacking in modern training enhancements such as backpropagation. KANs, however, allow the stacking of KAN-layers to create deep learning networks that can be trained using backpropagation.

Our work extends the foundational concepts of KANs to the realm of graph-based data, introducing Graph Kolmogorov-Arnold Networks (GKANs). These networks are designed to overcome the scalability and adaptability limitations faced by conventional GCNs. Similar to KANs, by incorporating learnable functions directly within the graph's edges, GKANs allow for a more dynamic feature propagation across expansive and evolving graph structures. We are motivated to investigate whether this approach holds the promise of surpassing the current graph learning paradigms. 

Through software experiments\footnote{We plan to make our software implementation of GKAN publicly available soon on Github at the following URL: \url{https://github.com/ANRGUSC/GKAN}.
} grounded in a real dataset, we find that GKANs significantly outperform state-of-the-art GCNs in terms of classification accuracy and efficiency. 
Our contribution lies in harmoniously blending the learning the graph structure of data with the innovative architecture of KANs, thereby setting a new standard for graph-based deep learning models to handle large-scale, dynamic datasets effectively. Though we focus our attention in this paper on GCNs, we believe that GKANs open a new avenue in graph representation learning and could serve as the foundation for all kinds of approaches that previously utilized MLPs at their core, such as GCNs [\cite{Hamilton}, \cite{Kipf}, \cite{Monti}, \cite{Rianne}, \cite{pinsage}], GAT [\cite{GAT},\cite{GAT_how_attentive}], Graph Autoencoders [\cite{Graph_AutoEncoder_Kipf},\cite{Graph_AutoEncoder_One-Hop}], Graph Transformers [\cite{Graph_Transformer},\cite{Graph_Transformer_Graph-to-Sequence},\cite{Graph_Transformer_GraphiT},\cite{Graph_Transformer_structure_aware}], and many other graph deep learning schemes.

In the following section, we will first provide a brief overview on KAN. Then we elaborate upon the GKAN architecture, and present empirical evidence of its superiority in node classification tasks.

\section{Preliminary}

\subsection{Kolmogorov–Arnold Networks}
KANs are inspired by the mathematical underpinning provided by the famous Kolmogorov-Arnold representation theorem~[\cite{KA_theoren_2},\cite{KA_theoren_3}]. This theorem offers a compelling framework for constructing neural networks that dramatically diverge from traditional Multi-Layer Perceptrons (MLPs), which are typically grounded in and justified by the universal approximation theorem. The unique approach of KANs utilizes a network design where traditional weight parameters are replaced by learnable functions. In Section 2.1, we review the origins and implications of the Kolmogorov-Arnold theorem, setting the stage for a deep dive into the innovative structure of KANs outlined in Section 2.2. 

\subsubsection{Kolmogorov-Arnold Theorem}
Through a series of papers in the late 1950's~[\cite{KA_theoren_1},\cite{KA_theoren_2},\cite{arnold1957functions},\cite{arnold1958representation},\cite{arnol1959representation}], Andrey Kolmogorov and Vladimir Arnold formulated a significant theorem which posits that any continuous multivariate function within a bounded domain can be effectively broken down into a series of continuous univariate functions combined through the operation of addition. 

The Kolmogorov-Arnold Representation theorem~[\cite{KA_theoren_3}] states that a smooth function \( f: [0, 1]^n \to \mathbb{R} \) can be expressed as:
\begin{equation}
    f(x_1, \ldots, x_n) = \sum_{q=1}^{2n+1} \Phi_q \left( \sum_{p=1}^n \phi_{q,p}(x_p) \right),
\end{equation}
where each \( \phi_{q,p} \) is a mapping from \([0, 1]\) to \(\mathbb{R}\), and each \( \Phi_q \) is a real-valued function. This formulation demonstrates that multivariate functions can fundamentally be reduced to a suitably defined composition of univariate functions, where the composition only involves simple addition.

Early work~\cite{girosi1989representation} argued that despite its elegance and generality, this theorem is not useful for machine learning because the inner-functions $\phi_{q,p}(.)$ are not in general smooth. Indicating that there were prior works on this topic that were not particularly successful in spurring wider interest and adoption, \cite{KAN} maintain a more positive view regarding the applicability of the Kolmogorov-Arnold theorem in machine learning contexts. Firstly, compared to other prior works that used this theorem in a limited way, they do not constrain the neural network to adhere to the original Equation (2.1), characterized by its shallow structure of two-layer nonlinearities and limited number of terms (2n + 1) in the hidden layer; instead, they propose extending the network to arbitrary widths and depths. Secondly, the functions commonly encountered in scientific and everyday contexts tend to be smooth and exhibit sparse compositional structures, which may enhance the effectiveness of Kolmogorov-Arnold representations.

\subsubsection{KAN Architecture}
In the domain of supervised learning tasks, where we deal with input-output pairs \(\{(x_i, y_i)\}\), their goal is to model a function \(f\) such that \(y_i \approx f(x_i)\) for all data points. Drawing inspiration from the structure proposed in Equation (2.1), ~\cite{KAN} design a neural network that explicitly embodies this equation. This network is constructed such that all learnable functions are univariate, with each being parameterized as a B-spline, enhancing the flexibility and learnability of the model.

Liu et al.'s initial model of the KAN embodies Equation (2.1) within its computation graph, showcasing a two-layer neural network where activation functions are located on edges rather than nodes, and aggregation is achieved through simple summation at nodes. This setup is depicted with an input dimensionality of \(n=2\) and a middle layer width of \(2n+1\), as illustrated in their figures.

Given the simplistic nature of this initial model, it is generally insufficient for approximating complex functions with high fidelity. To overcome this, Liu et al. propose an expansion of the KAN structure to include multiple layers, thereby increasing both its depth and breadth. The analogy to Multi-Layer Perceptrons (MLPs) becomes apparent here, as just like in MLPs, once a basic layer structure is defined—comprising linear transformations and nonlinearities—Liu et al. extend the model by adding more layers.

A KAN layer, suitable for a deeper architecture, is defined by a matrix of univariate functions:
\[
\Phi = \{\phi_{q,p}\}, \quad p = 1, 2, \ldots, n_{\text{in}}, \quad q = 1, 2, \ldots, n_{\text{out}},
\]
where each function \(\phi_{q,p}\) has trainable parameters. This enables the Kolmogorov-Arnold representations, initially described as a composition of two such layers, to be expanded into deeper configurations.

For a concrete and intuitive understanding of the KAN architecture, consider its representation as an array of integers \([n_0, n_1, \ldots, n_L]\), where \(n_i\) denotes the number of nodes in the \(i\)-th layer. Activation values and their transformations between layers are defined as follows:
\begin{equation}\label{eq:kan_layer}
x_{l+1,j} = \sum_{i=1}^{n_l} \phi_{l,j,i}(x_{l,i}),    
\end{equation}
where $\phi_{l,j,i}$ connects node \(i\) of layer $l$ with node $j$ of layer $l+1$. This setup allows Liu et al. to stack deep KAN layers, facilitating the modeling of complex functions through successive transformations and summations.

\subsection{Graph Convolutional Networks}
The conventional GCNs assume that node labels ${\bf y}$ are function of both graph structure  (i.e., adjacency matrix $A$) and node features $X$, or more formally speaking ${\bf y}=f(X,A)$. A multi-layer Graph Convolutional Network (GCN) is characterized by the layer-wise propagation rule:
\begin{equation}
    H^{(l+1)} = \sigma \left( \tilde{D}^{-\frac{1}{2}} \tilde{A} \tilde{D}^{-\frac{1}{2}} H^{(l)} W^{(l)} \right),
\end{equation}
where $\tilde{A} = A + I_N$ is the adjacency matrix of the graph augmented with self-connections, $I_N$ is the identity matrix, and $\tilde{D}_{ii} = \sum_j \tilde{A}_{ij}$. Here, $W^{(l)}$ represents the trainable weight matrix at layer $l$, and $\sigma(\cdot)$ denotes an activation function such as the ReLU function. The matrix \(H^{(l)} \in \mathbb{R}^{N \times D}\) encapsulates the activations in the $l$-th layer, with $H^{(0)} = X$.

In \cite{Kipf}, they showed that the above propagation rule is motivated by a first-order approximation of localized spectral filters on graphs, a concept pioneered by \cite{HAMMOND} and further explored by \cite{Defferrard}. The application of this rule involves iterative transformations across the graph structure, incorporating both node features and the topology of the graph to learn useful hidden layer representations in a highly scalable manner.

\begin{figure}[h]
    \centering
    \begin{tabular}{|m{.9cm}|m{0.8\textwidth}|}
        \hline
        \textbf{Scheme} & \textbf{Overview of Model Architecture} \\
        \hline
        \rotatebox{90}{GCN~\cite{pinsage}~~~~} & 
        \begin{subfigure}{.8\textwidth}
            \includegraphics[width=\linewidth, trim={0mm 20mm 0mm 40mm}, clip]{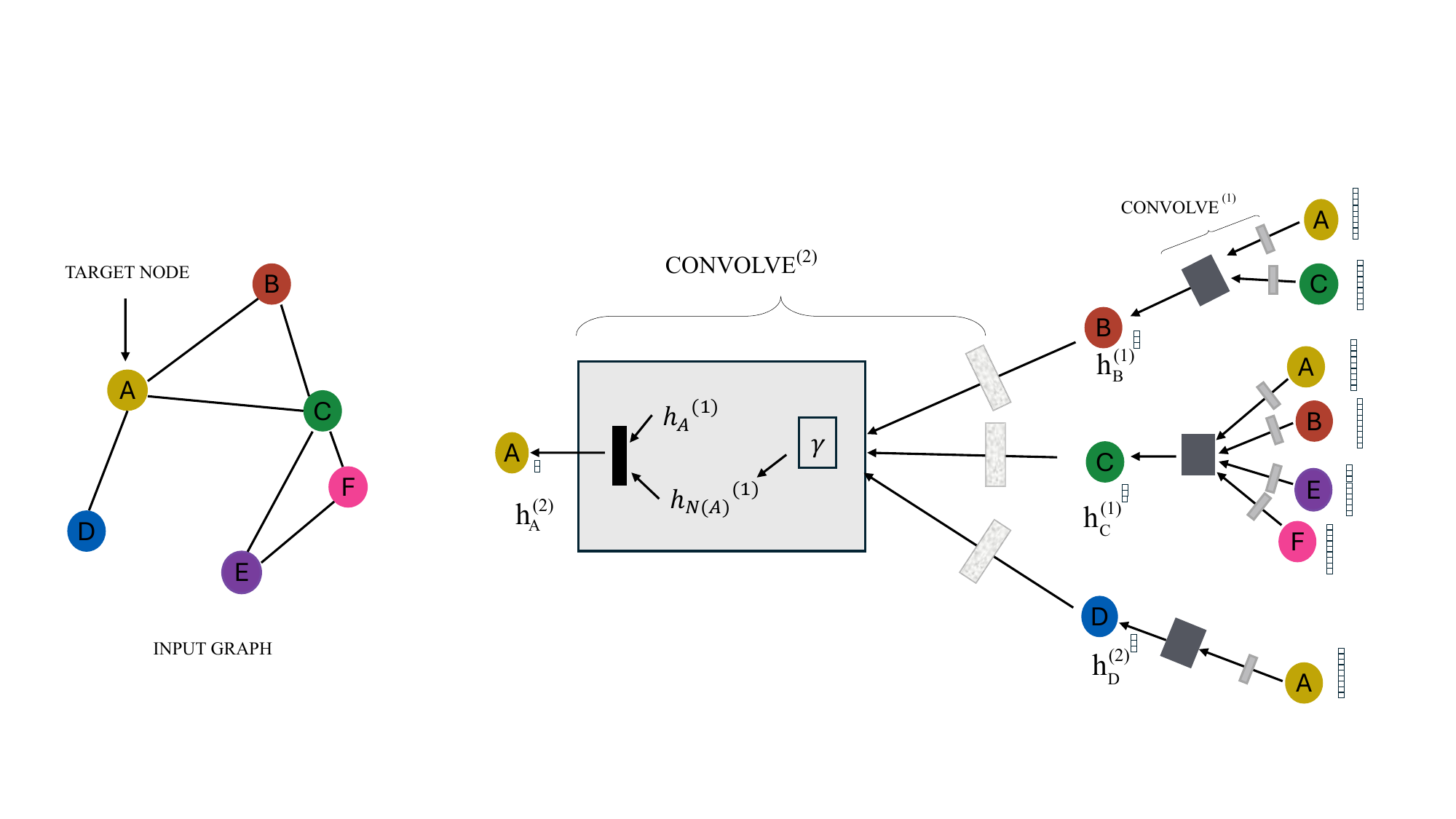} 
            \caption{Overview of a two-layer GCN~\cite{pinsage} architecture.}
        \end{subfigure} \\
        \hline
        \rotatebox{90}{GKAN Architecture 1~~~~~~~~} & 
        \begin{subfigure}{.8\textwidth}
            \includegraphics[width=\linewidth, trim={0mm 35mm 0mm 15mm}, clip]{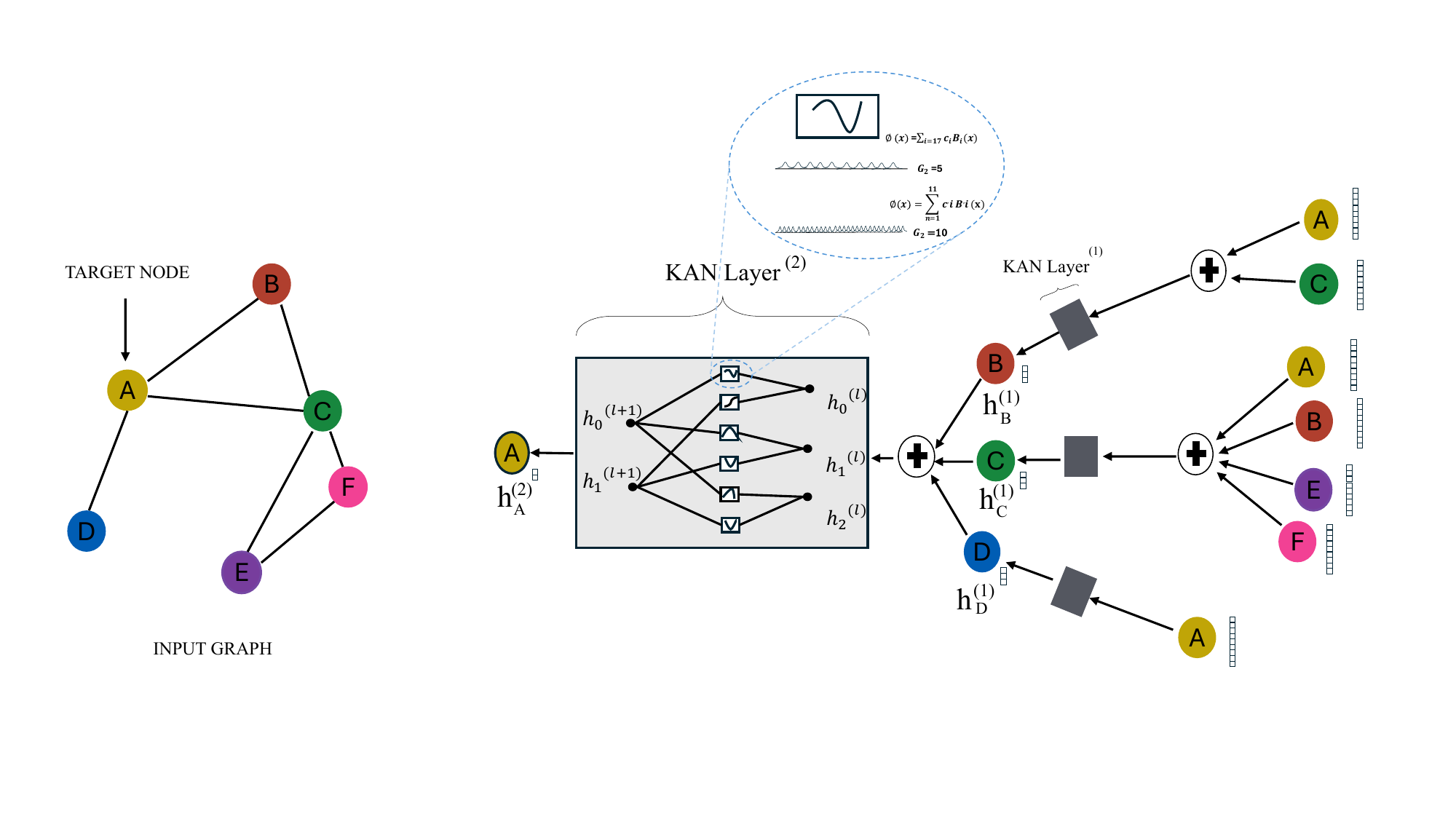}
            \caption{Overview of a two-layer GKAN Architecture 1.}
        \end{subfigure} \\
        \hline
        \rotatebox{90}{GKAN Architecture 2~~~~~~~~} & 
        \begin{subfigure}{.8\textwidth}
            \includegraphics[width=\linewidth, trim={0mm 20mm 0mm 40mm}, clip]{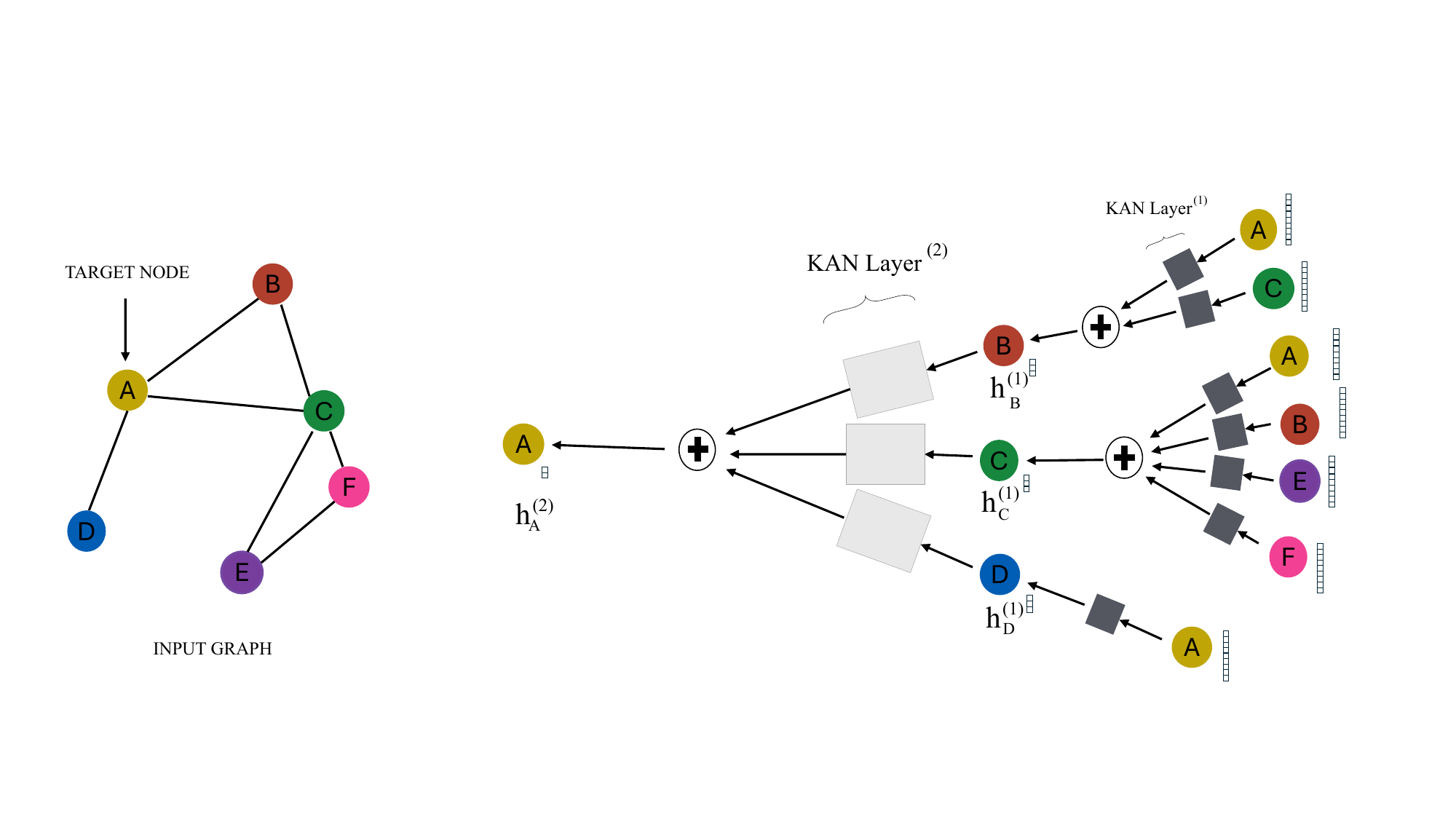}
            \caption{Overview of a two-layer GKAN Architecture 2.}
        \end{subfigure} \\
        \hline
    \end{tabular}
    \caption{Comparison of different model architectures.}\label{fig:Model_Architectures}
\end{figure}

For the sake of illustration, we show a conventional two-layer GCN of a graph node classification with a symmetric adjacency matrix, whether binary or weighted, in Figure 1. We can perform a preprocessing step by computing $\hat{A} = \tilde{D}^{-\frac{1}{2}} \tilde{A} \tilde{D}^{-\frac{1}{2}}$. Then, the forward model is expressed succinctly as:
\begin{equation}
Z = f(X,A) = \text{softmax}\left( \hat{A} \text{ReLU}\left( \hat{A} X W^{(0)} \right) W^{(1)} \right),
\end{equation}
where $W^{(0)} \in \mathbb{R}^{C \times H}$ represents the weight matrix that maps inputs to a hidden layer with $H$ feature maps, while $W^{(1)} \in \mathbb{R}^{H \times F}$ maps hidden layer features to the output layer. The softmax activation function is applied to each row and is defined as $\text{softmax}(x_i) = \frac{\exp(x_i)}{\sum_i \exp(x_i)}$. For tasks involving semi-supervised multi-class classification, the cross-entropy error is calculated across all labeled examples by:
$L = -\sum_{l \in Y_L} \sum_{f=1}^F Y_{lf} \ln Z_{lf}$,
where $Y_L$ denotes the set of node indices that are labeled. The weights of the neural network, $W^{(0)}$ and $W^{(1)}$, are optimized using the gradient descent method. The spatial-based graph representation of the embeddings described above is presented in Fig. \ref{fig:Model_Architectures}(a), as introduced by \cite{pinsage}.

\section{Proposed GKAN Achitectures}

In this model, the process of node classification leverages both the graph structure and node features through successive layers of transformation and non-linearity, providing a powerful mechanism for learning from both labeled and unlabeled data in a semi-supervised setting. The elegance of this approach lies in its simplicity and effectiveness, enabling the model to learn complex patterns of node connectivity and feature distribution efficiently.

We propose two simple architectures of GKAN, named Architecture 1 and Architecture 2, which we elaborate upon in the following subsections.
\subsection{GKAN Architecture 1}
In this architecture, the embedding of nodes at layer $\ell+1$ are basically generated by passing the aggregated (e.g., summation) node embedding at layer $\ell$ through $\text{KANLayer}^{(\ell)}$. More mathematically, 
\begin{align}
    H^{(\ell+1)}_{\text{Archit. 1}} = \text{KANLayer}(\hat{A}H^{(\ell)}_{\text{Archit. 1}})
\end{align}
with $H^{(0)}_{\text{Archit. 1}} = X$. 
The implementation of KANLayer is presented in Appendix.
Considering $L$ number of layers for this architectures, then the forward model is presented as $Z = \text{softmax}(H^{(L)}_{\text{Archit. 1}})$. This is a spectral-based representation, the construction of embedding of nodes in different layers of spatial-based representation of GKAN Architecture 1 is presented in Fig. \ref{fig:Model_Architectures}(b).

\subsection{GKAN Architecture 2}
Instead of performing the aggregation (according to the normalized adjacency matrix) and passing the aggregation through the KANLayer, we first pass the embedding through the KANLayer, then aggregating the result according to the normalized adjacency matrix. Formally speaking, the embedding of nodes at layer $\ell+1$ are as follows, 
\begin{align}
    H^{(\ell+1)}_{\text{Archit. 2}} = \hat{A}~\text{KANLayer}(H^{(\ell)}_{\text{Archit. 2}})
\end{align}
with $H^{(0)}_{\text{Archit. 2}} = X$.
Similar to the first architecture, the forward model would be $Z = \text{softmax}(H^{(L)}_{\text{Archit. 2}})$. Similarly, this is a spectral-based representation of the second architecture. The construction of embedding of nodes in different layers of spatial-based representation of GKAN Architecture 2 is presented in Fig. \ref{fig:Model_Architectures}(c).

\section{Experiments}

\textbf{Data:} We consider the Cora dataset, a citation network described in ~\cite{Sen_Namata_Bilgic_Getoor_Galligher_Eliassi}, consists of 2,708 nodes and 5,429 edges where nodes represent documents and edges signify citation links. It encompasses 7 different classes and includes 1,433 features per document. The distribution of the dataset is as follows: the training set contains 1000 samples, the validation set contains 200 samples, and the test set includes 300 samples.
Cora features sparse bag-of-words vectors for each document, and the citation links among these documents are treated as undirected edges to construct a symmetric adjacency matrix. Each document has a class label, and although only 20 labels per class are actively used for training, all feature vectors are incorporated. These details underscore the structured and characteristic nature of the Cora dataset, which is crucial for training and testing within the network.

\textbf{Overview of Experiments:} We present our experiments in two parts. First, we compare the accuracy we obtain with the two GKAN architectures with the accuracy obtained by the conventional GCN, over both train and test data. One subtle but important issue that we take into account when comparing GCN with GKAN on a fair-basis is to ensure that the parameter sizes are the same. Second, we examine the impact of various parameters for GKANs such as number of hidden nodes and the order and grid parameters for the B-splines used for the learnable univariate functions. 

\subsection{Comparison with GCN}
In order to fairly evaluate the performance of GKAN compared to GCN, we ensure that both networks have an identical number of layers with the dimensions specified as 
\begin{align}\nonumber
\text{GCN} &:[d_{\text{input}}:h_{GCN}:C]\\
\text{GKAN Architecture 1} &:[d_{\text{input}}:h:C]\\
\text{GKAN Architecture 2} &:[d_{\text{input}}:h:C]
\end{align}
where $d_{\text{input}}$, $h_{GCN}$, $h$, and $C$ respectively represent the dimension of input features, the dimension of hidden layers of GCN, the dimension of hidden layers of GKAN architectures, and the total number of classes. 
To ensure a fair comparison between GKAN and GCN, we equipped GCN with a higher $h_{GCN}$ compared to $h$
to approximately equalize the total number of parameters in the two schemes.
The total number of trainable parameters of GCN and GKAN Architectures 1 and GKAN Architecture 2 are measured by the parameter counting module in PyTorch.

The accuracy of our proposed GKAN for different settings of \(k\) (the degree of the polynomial in KAN settings) and \(g\) (the number of intervals in KAN settings) against GCN~\cite{Kipf} on Cora dataset using the first 100 features is shown in Table~\ref{tab:architecture_performance_cora_100}.
We further present the the accuracy of GKAN architectures compared to GCN~\cite{Kipf} on Cora dataset using the first 200 features in Table~\ref{tab:architecture_performance_cora_200}. The results in these two tables show that all the GKAN variants considered perform better in terms of accuracy compared to a GCN with compable number of parameters. Further, they suggest that architecture 2 generally performs better. For the case of 100 features, the best GKAN model shows more than 8\% higher accuracy than the conventional GCN model; for 200 features, the best GKAN model show a more than 6\% higher accuracy than the conventional GCN model.

\begin{table}[ht]
\centering
\caption{Architectures and their performance on the first 100 features of Cora Dataset.}
\label{tab:architecture_performance_cora_100}
\begin{tabular}{ccc} 
\toprule
\textbf{Architecture} & \textbf{\#Parameters} & \textbf{Test} \\
\midrule
$\text{GCN}_{h_{GCN}=205}$ & 22,147 & 53.50 \\
$\text{GKAN}_{(k=1, g=10, h=16)}^{\text{(Archit. 1)}}$ & 22,279 & 59.32 \\
$\text{GKAN}_{(k=1, g=10, h=16)}^{\text{(Archit. 2)}}$ & 22,279 & 61.48 \\
$\text{GKAN}_{(k=2, g=9, h=16)}^{\text{(Archit. 1)}}$ & 22,279 & 56.76 \\
$\text{GKAN}_{(k=2, g=9, h=16)}^{\text{(Archit. 2)}}$ & 22,279 & {\bf 61.76} \\
\bottomrule
\end{tabular}
\end{table}

\begin{table}[ht]
\centering
\caption{Architectures and their performance on the first 200 features of Cora Dataset.}
\label{tab:architecture_performance_cora_200}
\begin{tabular}{ccc} 
\toprule
\textbf{Architecture} & \textbf{\#Parameters} & \textbf{Test} \\
\midrule
$\text{GCN}_{h_{GCN}=104}$ & 21,639 & 61.24 \\
$\text{GKAN}_{(k=2, g=2, h=17)}^{\text{(Archit. 1)}}$ & 21,138 & 63.58 \\
$\text{GKAN}_{(k=2, g=2, h=17)}^{\text{(Archit. 2)}}$ & 21,138 & 64.10 \\
$\text{GKAN}_{(k=1, g=2, h=20)}^{\text{(Archit. 1)}}$ & 20,727 & 67.44 \\
$\text{GKAN}_{(k=1, g=2, h=20)}^{\text{(Archit. 2)}}$ & 20,727 & {\bf 67.66} \\
\bottomrule
\end{tabular}
\end{table}

In the following parts, we measure the performance of GCN as well as the proposed GKAN architectures 1 and 2 on Cora dataset considering only the first 100 features.

\subsubsection{Accuracy and Loss Values vs. Epochs}
We set \(h = 16\) and \(h_{GCN} = 100\) to have a fair comparison between $\text{GKAN}_{(k=1,g=3,h=16)}$ and $\text{GCN}_{(h_{GCN}=100)}$ ensuring almost the same number of trainable parameters. 
For this settings, the total number of trainable parameters of GCN, GKAN Architectures 1 and GKAN Architecture 2 are $10807$, $10295$, and $10295$, respectively. 

\begin{figure}[htbp]
\centering
\begin{subfigure}{0.5\textwidth}
    \centering
    \includegraphics[scale=0.4]{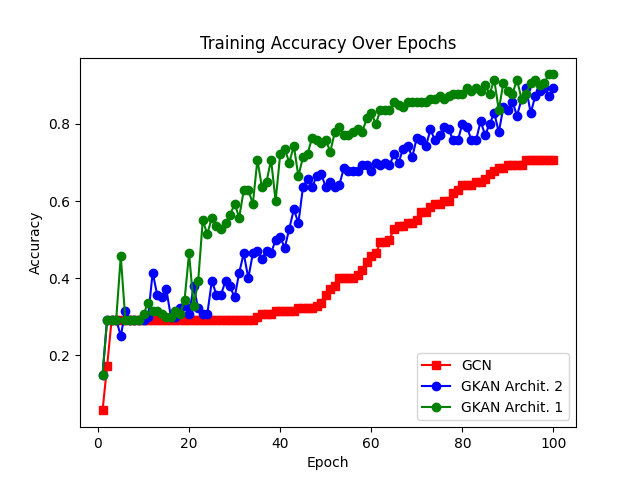}
    \caption{Training accuracy of different schemes.}
    \label{plot:train_acc_vs_epoch_k1_g3}
\end{subfigure}%
\begin{subfigure}{0.5\textwidth}
    \centering
    \includegraphics[scale=0.4]{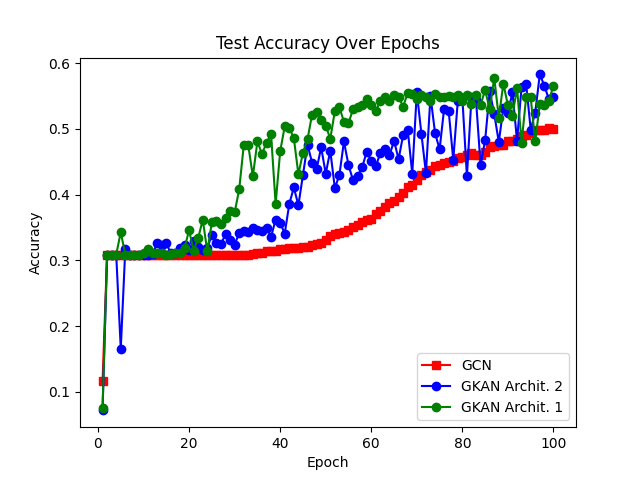}
    \caption{Test accuracy of different schemes.}
    \label{plot:test_acc_vs_epoch_k1_g3}
\end{subfigure}
\caption{Accuracy comparison of GCN and GKAN architectures for $k=1$ and $g=3$.}
\label{plot:acc_vs_epoch_k1_g3}
\end{figure}

\begin{figure}[htbp]
\centering
\begin{subfigure}{0.5\textwidth}
    \centering
    \includegraphics[scale=0.4]{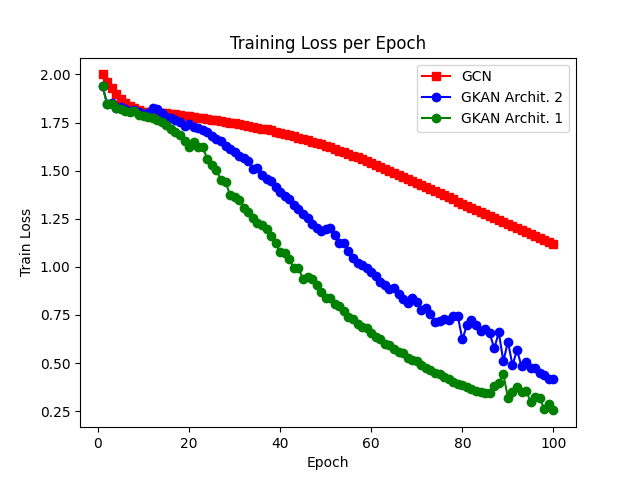}
    \caption{Training loss of different schemes.}
    \label{plot:train_loss_vs_epoch_k1_g3}
\end{subfigure}%
\begin{subfigure}{0.5\textwidth}
    \centering
    \includegraphics[scale=0.4]{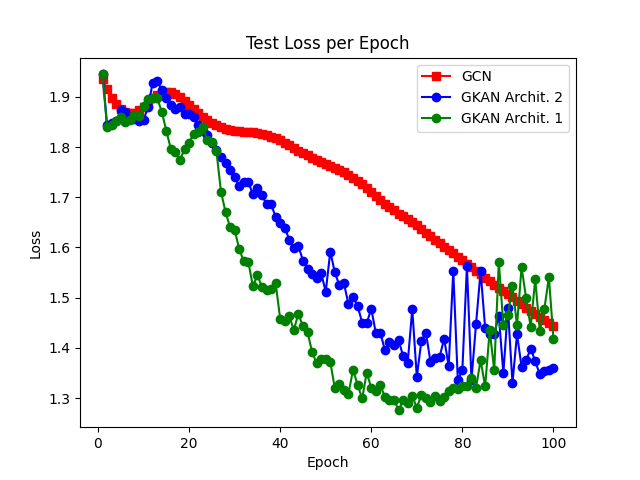}
    \caption{Test loss of different schemes.}
    \label{plot:test_loss_vs_epoch_k1_g3}
\end{subfigure}
\caption{Loss value comparison of GCN and GKAN architectures for $k=1$ and $g=3$.}
\label{plot:loss_vs_epoch_k1_g3}
\end{figure}

Figures \ref{plot:train_acc_vs_epoch_k1_g3} and \ref{plot:test_acc_vs_epoch_k1_g3} represent the accuracy of training and test for the GCN, and our proposed GKAN architectures with $k=1$ and $g=3$ on Cora dataset with the first $100$ features. As it can be observed, GKAN architectures improves test accuracy of GCN by a large margin of $6\%$. 
Figures \ref{plot:train_loss_vs_epoch_k1_g3} and \ref{plot:test_loss_vs_epoch_k1_g3} illustrate training loss and test loss for the GCN as well as our proposed GKAN architectures for the same settings of $k=1$ and $g=3$. As one can see, GKAN architectures lead to a sharp decrease in the loss values, implying that the GKAN architecture requires less number of epochs to be trained, matching previous observations~\cite{KAN} about KAN compared to Multi-Layer Perceptron (MLP).  

\subsection{Evaluating the Influence of Parameters on GKAN}

In this section, we explore the impact of the parameters $g$ (spline grid size), $k$ (spline polynomial degree), and $h$ (number of hidden nodes) on the performance of GKAN. To isolate the influence of each parameter, we vary one while holding the others at their default values. The default settings for $g$, $k$, and $h$ are 3, 1, and 16, respectively. Table \ref{tab:parameter_values_range} outlines the range of values tested for each parameter, providing a structured framework for our investigation. This approach allows us to systematically determine how each parameter affects the overall effectiveness of our model. Given our earlier observations, we focus on GKAN architecture 2 in this section. 

\begin{table}[h]
\centering
\caption{Range of Values for Parameters for the degree of the spline polynomial \( k \), grid size for spline \( g \), and number of hidden nodes \( h \), with default values in bold}
\label{tab:parameter_values_range}
\begin{tabular}{cc}
\toprule
Parameter & Values \\
\midrule
\( k \) & \{\textbf{1}, 2, 3\} \\
\( g \) & \{\textbf{3}, 7, 11\} \\
\( h \) & \{8, 12, \textbf{16}\} \\
\bottomrule
\end{tabular}
\end{table}

\subsubsection{Effect of varying grid size $g$}
\begin{figure}[htbp]
\centering
\begin{subfigure}{0.45\textwidth}
    \centering
    \includegraphics[scale=0.4]{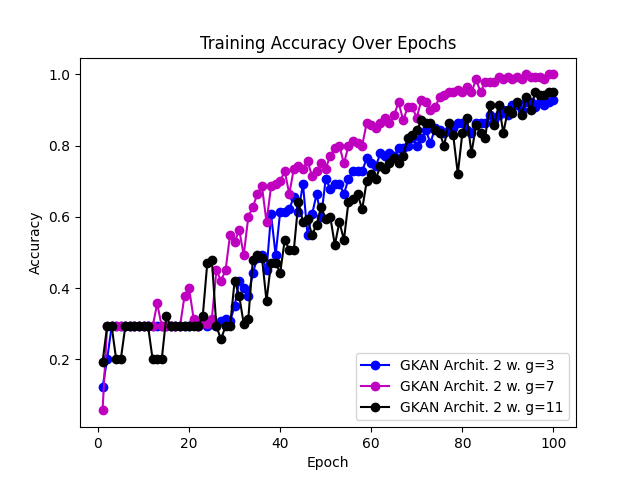}
    \caption{Training accuracy of GKAN Architecture 2 for different parameter $g$ and $k=1$.}
    \label{plot:GRID_train_acc_vs_epoch_k1}
\end{subfigure}%
\hspace{2mm}
\begin{subfigure}{0.45\textwidth}
    \centering
    \includegraphics[scale=0.4]{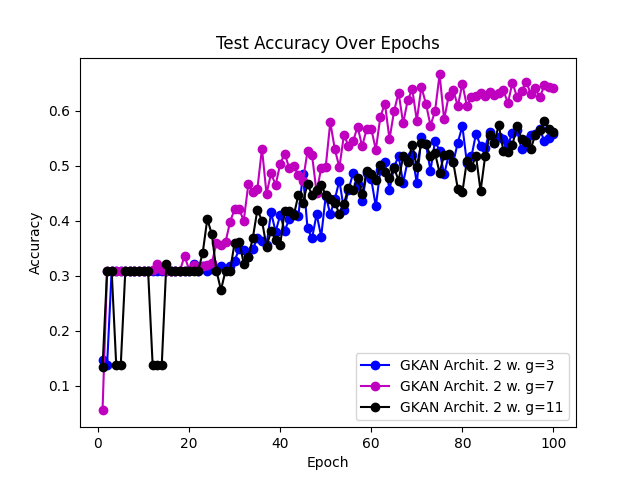}
    \caption{Test accuracy of GKAN Architecture 2 for different parameter $g$ and $k=1$.}
    \label{plot:GRID_test_acc_vs_epoch_k1}
\end{subfigure}
\caption{Accuracy comparison of GKAN Architecture 2 for $g\in\{1,2,3\}$, $k=1$, and $h=16$.}
\label{plot:GRID_acc_vs_epoch_k1}
\end{figure}

Figures \ref{plot:GRID_train_acc_vs_epoch_k1} and \ref{plot:GRID_test_acc_vs_epoch_k1} show the accuracy of GKAN Architecture 2 for different values of grid $g$ (i.e., $g=3$, $g=7$, and$g=11$) and fixed $k=1$. As one can see, based on the validation performance, the best choice of $g$ for this problem among these three is an intermediate value of $g=7$. It appears that while there is some benefit to increasing the grid size from $g=3$ to $g=7$, a higher value of $g=11$ causes a deterioration and results in a performance that is comparable to $g=3$.

\begin{figure}[htbp]
\centering
\begin{subfigure}{0.45\textwidth}
    \centering
    \includegraphics[scale=0.4]{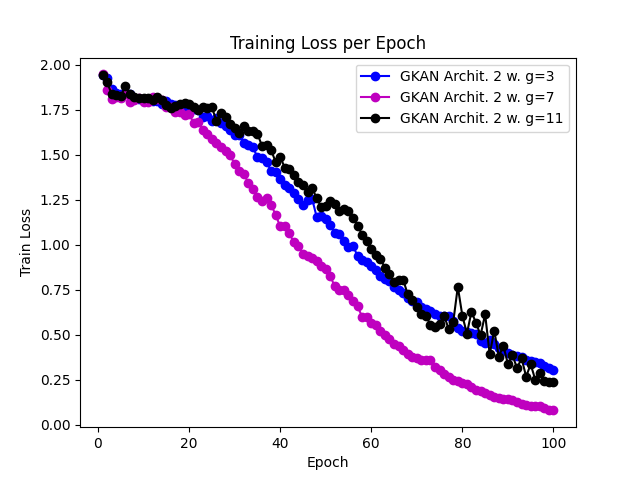}
    \caption{Training loss values of GKAN Architecture 2 for different parameter $g$ and $k=1$.}
    \label{plot:GRID_train_loss_vs_epoch_k1}
\end{subfigure}%
\hspace{2mm}
\begin{subfigure}{0.45\textwidth}
    \centering
    \includegraphics[scale=0.4]{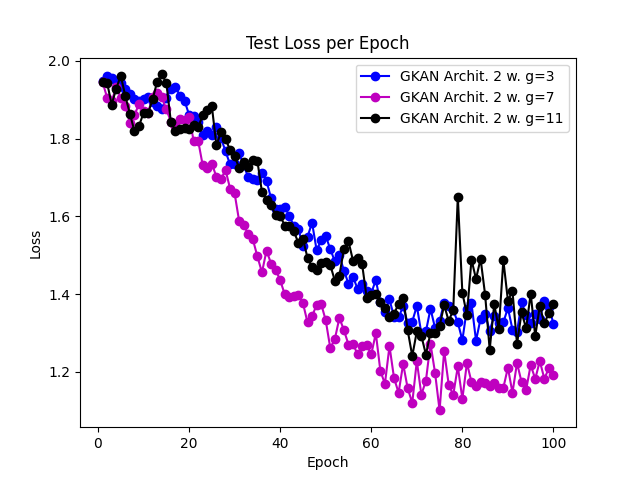}
    \caption{Test loss values of GKAN Architecture 2 for different parameter $g$ and $k=1$.}
    \label{plot:GRID_test_loss_vs_epoch_k1}
\end{subfigure}
\caption{Loss value of GKAN Architecture 2 for $g\in \{3,7,11\}$ and $k=1$, and $h=16$.}
\label{plot:GRID_loss_vs_epoch_k1}
\end{figure}

The loss values of training and test for GKAN Architecture 2 for different values 
 of grid parameter $g$ and fixed degree $k=1$ are presented in figures \ref{plot:GRID_train_loss_vs_epoch_k1} and \ref{plot:GRID_test_loss_vs_epoch_k1}, respectively.

\subsubsection{Effect of varying the degree of polynomials $k$}
\begin{figure}[htbp]
\centering
\begin{subfigure}{0.45\textwidth}
    \centering
    \includegraphics[scale=0.4]{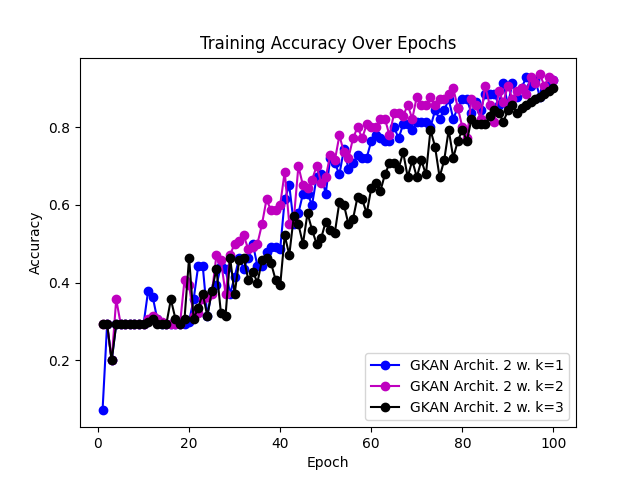}
    \caption{Training accuracy of GKAN Architecture 2 for $k\in\{1,2,3\}$ while fixing $g=3$.}
    \label{plot:DEGREE_train_acc_vs_epoch_k1}
\end{subfigure}%
\hspace{2mm}
\begin{subfigure}{0.45\textwidth}
    \centering
    \includegraphics[scale=0.4]{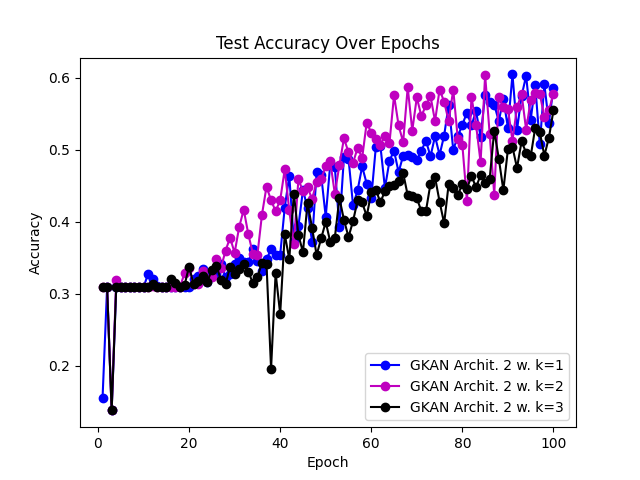}
    \caption{Test accuracy of GKAN Architecture 2 for $k\in\{1,2,3\}$ and fixed $g=3$.}
    \label{plot:DEGREE_test_acc_vs_epoch_k1}
\end{subfigure}
\caption{Accuracy of GKAN Architecture 2 for $k\in\{1,2,3\}$, $g=3$, and $h=16$.}
\label{plot:DEGREE_acc_vs_epoch_k1}
\end{figure}

\begin{figure}[htbp]
\centering
\begin{subfigure}{0.45\textwidth}
    \centering
    \includegraphics[scale=0.4]{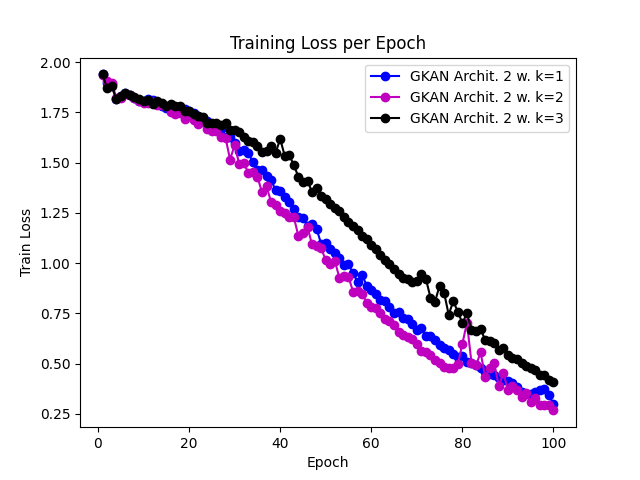}
    \caption{Training loss values of GKAN Architecture 2 for $k\in\{1,2,3\}$ and fixed $g=3$.}
    \label{plot:DEGREE_train_loss_vs_epoch_k1}
\end{subfigure}%
\hspace{2mm}
\begin{subfigure}{0.45\textwidth}
    \centering
    \includegraphics[scale=0.4]{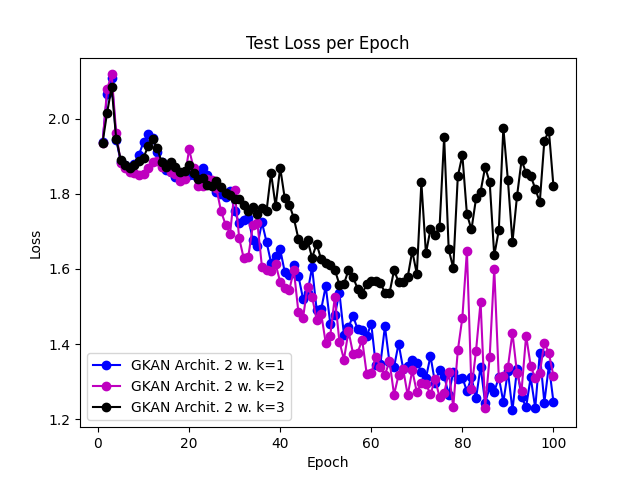}
    \caption{Test loss values of GKAN Architecture 2 for $k\in\{1,2,3\}$ and $g=3$.}
    \label{plot:DEGREE_test_loss_vs_epoch_k1}
\end{subfigure}
\caption{Loss value of GKAN Architecture 2 for $k\in\{1,2,3\}$, $g=3$ and $h=16$.}
\label{plot:DEGREE_loss_vs_epoch_k1}
\end{figure}

We present the accuracy of GKAN Architecture 2 for different values of degree $k$, ranging from $k=1$ to $k=3$ while fixing the grid size $g=3$ in figures 
\ref{plot:DEGREE_train_acc_vs_epoch_k1} and \ref{plot:DEGREE_test_acc_vs_epoch_k1}. 
We see that a degree value of one has the best performance among this range of $k$, suggesting the possibility of underlying ground-truth function to be piece-wise linear. Figures \ref{plot:DEGREE_train_loss_vs_epoch_k1} and \ref{plot:DEGREE_test_loss_vs_epoch_k1} illustrate training loss and test loss, respectively.

\subsubsection{Effect of varying the size of hidden layer}
\begin{figure}[htbp]
\centering
\begin{subfigure}{0.45\textwidth}
    \centering
    \includegraphics[scale=0.4]{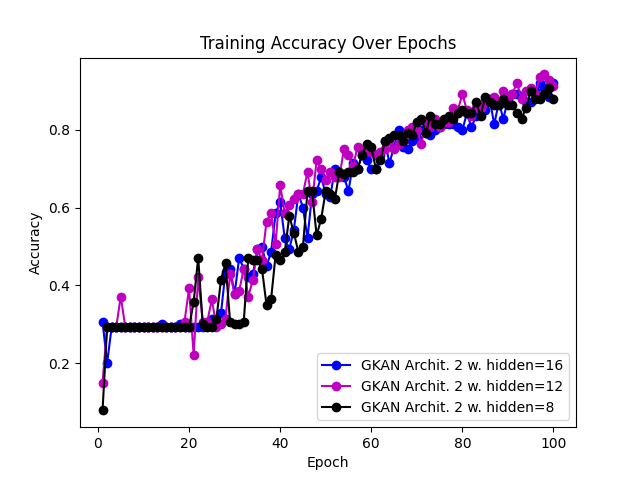}
    \caption{Training accuracy of GKAN Architecture 2 for $h\in\{8,12,16\}$ while fixing $g=3$ and $k=1$.}
    \label{plot:HIDDEN_train_acc_vs_epoch_k1}
\end{subfigure}%
\hspace{2mm}
\begin{subfigure}{0.45\textwidth}
    \centering
    \includegraphics[scale=0.4]{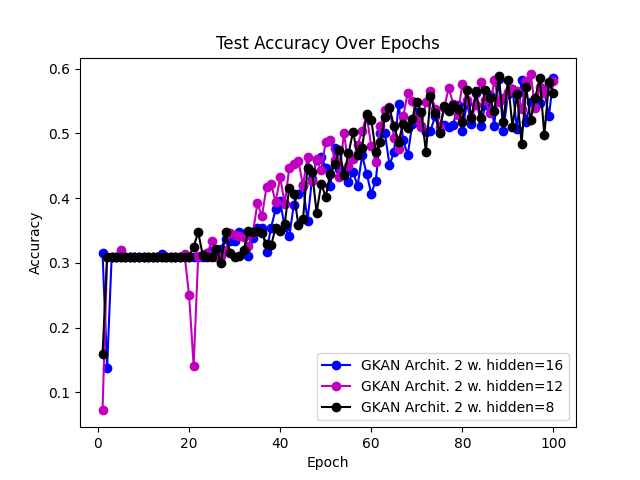}
    \caption{Test accuracy of GKAN Architecture 2 $h\in\{8,12,16\}$ while fixing $g=3$ and $k=1$.}
    \label{plot:HIDDEN_test_acc_vs_epoch_k1}
\end{subfigure}
\caption{Accuracy of GKAN Architecture 2 for $h\in\{8,12,16\}$, $g=3$ and $k=1$.}
\label{plot:HIDDEN_acc_vs_epoch_k1}
\end{figure}

\begin{figure}[htbp]
\centering
\begin{subfigure}{0.45\textwidth}
    \centering
    \includegraphics[scale=0.4]{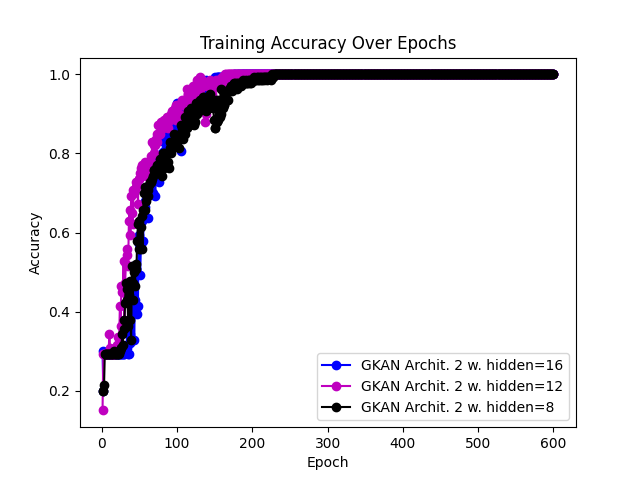}
    \caption{Training accuracy of GKAN Architecture 2 for $h\in\{8,12,16\}$ while fixing $g=3$ and $k=1$.}
    \label{plot:HIDDEN_train_acc_vs_epoch_k1_EPOCH_600}
\end{subfigure}%
\hspace{2mm}
\begin{subfigure}{0.45\textwidth}
    \centering
    \includegraphics[scale=0.4]{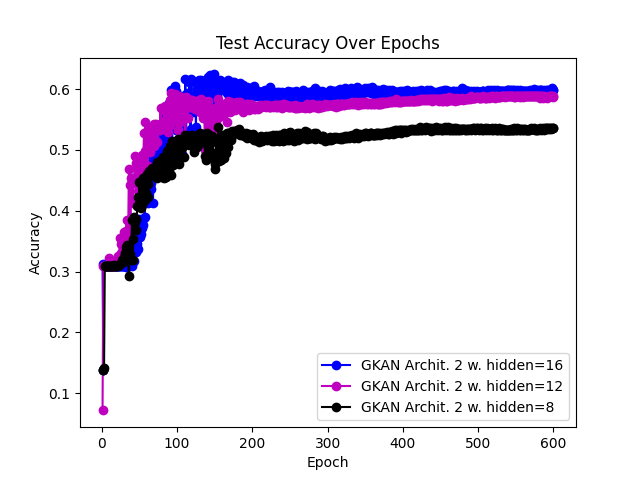}
    \caption{Test accuracy of GKAN Architecture 2 $h\in\{8,12,16\}$ while fixing $g=3$ and $k=1$.}
    \label{plot:HIDDEN_test_acc_vs_epoch_k1_EPOCH_600}
\end{subfigure}
\caption{Accuracy of GKAN Architecture 2 for $h\in\{8,12,16\}$, $g=3$ and $k=1$ over 600 epochs.}
\label{plot:HIDDEN_acc_vs_epoch_k1_EPOCH_600}
\end{figure}

Figures \ref{plot:HIDDEN_train_acc_vs_epoch_k1} and \ref{plot:HIDDEN_test_acc_vs_epoch_k1} illustrate the accuracy of GKAN Architecture 2 for the size of hidden layer $h\in\{8, 12, 16\}$ while fixing $k=1$ and $g=3$. Moreover, Figures \ref{plot:HIDDEN_train_acc_vs_epoch_k1_EPOCH_600} and \ref{plot:HIDDEN_test_acc_vs_epoch_k1_EPOCH_600}
demonstrate the accuracy of GKAN Architecture 2 for the same range of parameters over 600 epochs.
These results suggest that a hidden layer size of $h=12$ is particularly effective in the initial phases of training and ultimately achieves almost the same test performance as $h=16$.

\begin{figure}[htbp]
\centering
\begin{subfigure}{0.45\textwidth}
    \centering
    \includegraphics[scale=0.4]{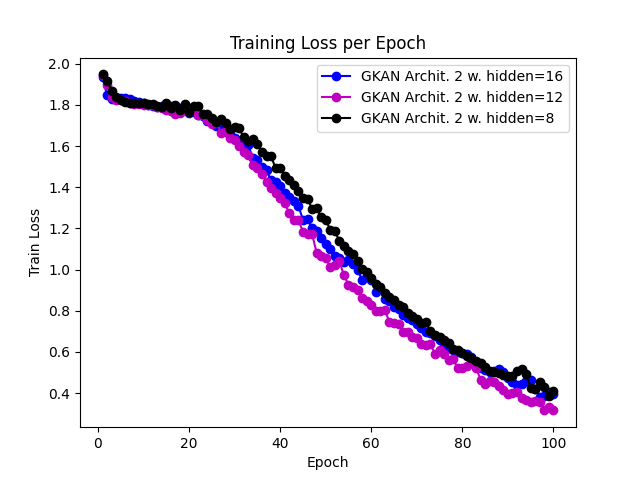}
    \caption{Training loss values of GKAN Architecture 2 for $h\in\{8,12,16\}$ while fixing $g=3$ and $k=1$.}
    \label{plot:HIDDEN_train_loss_vs_epoch_k1}
\end{subfigure}%
\hspace{2mm}
\begin{subfigure}{0.45\textwidth}
    \centering
    \includegraphics[scale=0.4]{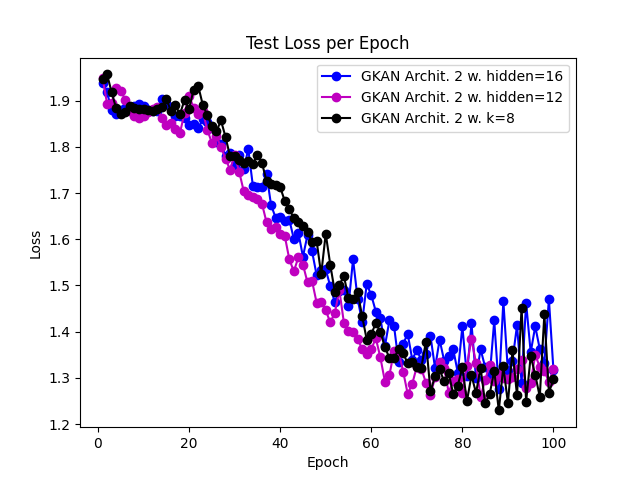}
    \caption{Test loss values of GKAN Architecture 2 for $h\in\{8,12,16\}$ while fixing $g=3$ and $k=1$.}
    \label{plot:HIDDEN_test_loss_vs_epoch_k1}
\end{subfigure}
\caption{Loss value of GKAN Architecture 2 for $h\in\{8,12,16\}$, $g=3$ and $k=1$.}
\label{plot:HIDDEN_loss_vs_epoch_k1}
\end{figure}

We also present the the loss values of training and test of GKAN Architecture 2 for $h\in\{8,12,16\}$ while fixing degree of polynomials to $k=1$ and grid size $g=3$ in figures \ref{plot:HIDDEN_train_loss_vs_epoch_k1} and \ref{plot:HIDDEN_test_loss_vs_epoch_k1}, respectively.

\section{Conclusions}

We considered how to apply the idea of learnable functions from the recently-proposed Kolmogorov-Arnold Neural Networks (KANs) to graph-structured data. In this work, we have presented, for the first time, two different architectures for Graph Kolmogorov-Arnold Networks (GKANs). Empirical evaluations on the Cora dataset show that GKANs attain significantly better parameter-efficiency than conventional GCN, yielding higher accuracy for comparable parameters sizes. We also examined how various parameters such as number of hidden nodes, grid size, and the spline order parameter impact performance.

Based on the evidence of the results presented in this paper, we believe that GKANs open a new avenue in graph representation learning and could serve as the foundation for all kinds of approaches that previously utilized MLPs at their core such as GCNs, GAT, Graph Autoencoders, Graph Transformers and many other graph deep learning schemes. Promising avenues for future work include exploring and evaluating extensions based on all these approaches using KAN, over more comprehensive datasets. GKAN currently inherit the property of present-generation KAN in that the training process is rather slow, and ~\cite{KAN} leave to future work the task of optimizing training time; advances in alternative learning approaches and architectures for KAN could also be applied to GKAN in the future.

\section*{Acknnowledgements:} 
This work was supported in part by Army Research Laboratory under Cooperative Agreement W911NF-17-2-0196. The authors acknowledge the Center for Advanced Research
Computing (CARC) at the University of Southern California
for providing computing resources that have contributed to
the research results reported within this publication. URL:
https://carc.usc.edu

\bibliographystyle{ACM-Reference-Format}
\bibliography{references}

\appendix

\section{KANLayer Implementation Details}
The KAN layer, although appearing simple in equation (\ref{eq:kan_layer}), presents challenges for optimization. The strategies employed by \cite{KAN} to overcome these challenges include:

\begin{enumerate}
    \item \textbf{Residual Activation Functions:} The activation function $\phi(x)$ combines a basis function $b(x)$, reminiscent of residual connections, and a spline function:
    \begin{equation}
        \phi(x) = w_b b(x) + w_s \text{spline}(x).
    \end{equation}
    The basis function is typically defined as:
    \begin{equation}
        b(x) = \text{silu}(x) = \frac{x}{1 + e^{-x}}.
    \end{equation}
    The spline component is expressed as a weighted sum of B-splines:
    \begin{equation}
        \text{spline}(x) = \sum_{i} c_i B_i(x),
    \end{equation}
    where $c_i$ are coefficients that can be adjusted during training. Notably, $w_b$ and $w_s$ could be absorbed into $b(x)$ and $\text{spline}(x)$ but are kept separate to fine-tune the function's amplitude.

    \item \textbf{Initialization Scales:} 
    The activation functions are initialized such that $w_s = 1$ and spline$(x) \approx 0$. $w_b$ is set based on the Xavier initialization scheme, traditionally used for initializing layers in MLPs.

    \item \textbf{Dynamic Spline Grid Updates:} The spline grids are updated dynamically based on the input activations. This modification caters to the inherent bounded nature of spline functions, accommodating the evolutionary nature of activation values during training.
\end{enumerate}

\end{document}